\documentclass[conference]{IEEEtran}
\IEEEoverridecommandlockouts
\usepackage{cite}
\usepackage{amsmath,amssymb,amsfonts}
\usepackage{algorithmic}
\usepackage{graphicx}
\usepackage{textcomp}
\usepackage{xcolor}
\usepackage{arydshln}
\usepackage{multirow}
\usepackage{multicol}

\def\BibTeX{{\rm B\kern-.05em{\sc i\kern-.025em b}\kern-.08em
    T\kern-.1667em\lower.7ex\hbox{E}\kern-.125emX}}
\begin{document}

\title{SemiSAM: Enhancing Semi-Supervised Medical Image Segmentation via SAM-Assisted Consistency Regularization}

\author{\IEEEauthorblockN{Yichi Zhang$^{1,2}$, Jin Yang$^{3}$, Yuchen Liu$^{1,2}$, Yuan Cheng$^{1,2}$, Yuan Qi$^{1,2}$} 
\IEEEauthorblockA{
\textit{$^{1}$ Artificial Intelligence Innovation and Incubation Institute, Fudan University, Shanghai, China} \\
\textit{$^{2}$ Shanghai Academy of Artificial Intelligence for Science, Shanghai, China}\\
\textit{$^{3}$ Department of Radiology, Washington University in St.Louis, MO, USA}}}

\maketitle

\begin{abstract}
Semi-supervised learning has attracted much attention due to its less dependence on acquiring abundant annotations from experts compared to fully supervised methods, which is especially important for medical image segmentation which typically requires intensive pixel/voxel-wise labeling by domain experts. Although semi-supervised methods can improve the performance by utilizing unlabeled data, there are still gaps between fully supervised methods under extremely limited annotation scenarios. 
In this paper, we propose a simple yet efficient strategy to explore the usage of the Segment Anything Model (SAM) for enhancing semi-supervised medical image segmentation. Concretely, the segmentation model trained with domain knowledge provides information for localization and generating input prompts to the SAM. Then the generated pseudo-labels of SAM are utilized as additional supervision to assist in the learning procedure of the semi-supervised framework. 
Extensive experiments demonstrate that SemiSAM significantly improves the performance of existing semi-supervised frameworks when only one or a few labeled images are available and shows strong efficiency as a plug-and-play strategy for semi-supervised medical image segmentation.
\end{abstract}


\section{Introduction}
Medical image segmentation aims to discern specific anatomical structures from medical images like organs and lesions, which is a basic and important step to provide reliable volumetric and shape information and assist in numerous clinical applications like disease diagnosis and quantitative analysis \cite{MSD,lalande2021deep,AbdomenCT-1K}.
Despite the remarkable performance of deep learning-based methods for medical image segmentation tasks, most of these methods require relatively large amounts of high-quality annotated data for training, while it is impractical to obtain large-scale carefully labeled datasets, particularly for medical imaging where only experts can provide reliable and accurate annotations \cite{tajbakhsh2020embracing}.
Besides, commonly used modalities of medical imaging like CT and MRI are 3D volumetric images, which further increase the manual annotation workload compared with 2D images where experts need to delineate from the volume slice by slice \cite{Zhang2020Bridging2A}.

To ease the manual labeling burden in response to this challenge, significant efforts have been devoted to annotation-efficient deep learning for medical image segmentation \cite{zhang2021exploiting,zhao2023one,zhu2021semi}.
Among these approaches, semi-supervised learning is a more practical method by encouraging models to utilize unlabeled data, which is much easier to acquire in conjunction with a limited amount of labeled data for training \cite{SemiSurvey}.
Semi-supervised learning is typically divided into two types: pseudo-label based method \cite{chen2021semi,thompson2022pseudo,zhao2023rcps} to assign pseudo labels for unlabeled data and train the model with both labeled and pseudo-labeled data, and consistency regularization-based methods \cite{luo2021semi,yu2019uncertainty,zhang2023uncertainty,zhang2021dual} to learn from both labeled and unlabeled data using unsupervised regularization. 
Although these methods can improve the performance by utilizing unlabeled data, there are still gaps between fully supervised methods under extremely limited annotation scenarios. When applying to new segmentation tasks, these semi-supervised methods still need to annotate a small subset of the dataset to obtain acceptable performance.

Recently, segmentation foundational models such as the Segment Anything Model (SAM) \cite{SAM-Meta} have attracted great attention with their powerful zero-shot generalization ability on a variety of semantic segmentation tasks \cite{semanticSAM,trackanything}. Although recent studies have revealed SAM’s limited performance in medical image segmentation due to the differences between natural images and medical images \cite{SAM-Empirical,SAM4MIS}, it still opened up new opportunities serving as a reliable pseudo-label generator to guide the segmentation task when manually annotated images are scarce \cite{li2023segment} by leveraging the knowledge base of foundation models.

In this paper, we propose a simple yet efficient strategy to explore the usage of SAM as an additional supervision branch for enhancing consistency regularization-based semi-supervised medical image segmentation framework especially in extremely limited annotation scenarios.
Specifically, the segmentation model provides information for localization and generating prompt points to the SAM.
Other than optimizing the segmentation model with supervised segmentation loss based on labeled cases and unsupervised consistency loss based on unlabeled cases, we leverage the consistency of predictions between SAM and the segmentation model as an additional supervision signal to assist in the learning procedure.
Extensive experiments on the Left Atrium (LA) dataset \cite{LAdataset} demonstrate that SemiSAM significantly enhances the segmentation performance particularly when only one or a few labeled images are available.

\begin{figure}[t]
	\includegraphics[width=9cm]{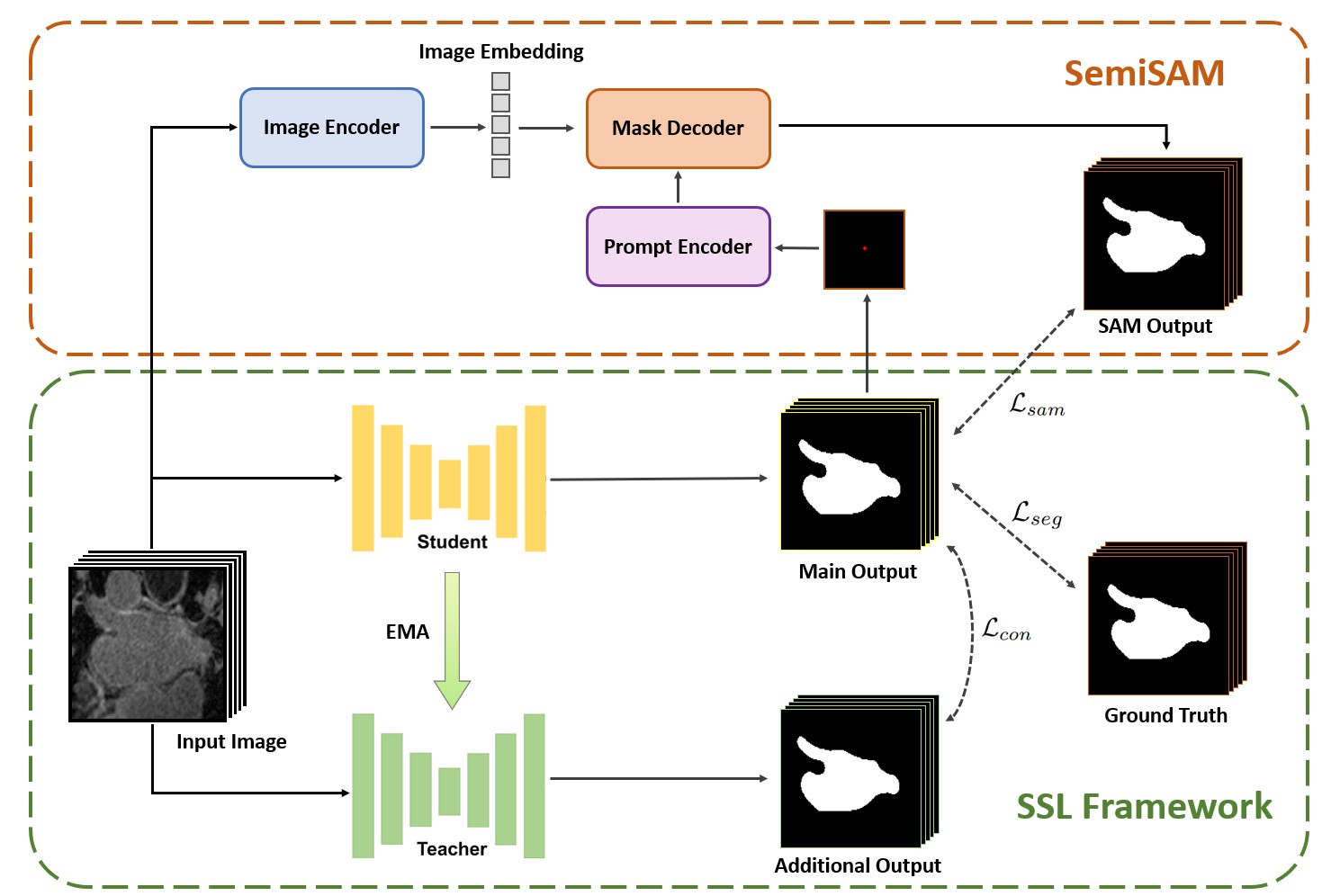}
	\caption{The structure of our proposed SAM-Assisted Consistency Regularization framework for semi-supervised medical image segmentation. The illustration in the figure is SemiSAM-MT based on the mean teacher framework \cite{tarvainen2017mean}.}
	\label{Architecture}
\end{figure}

\section{Methodology}

The structure of our proposed SAM-assisted consistency regularization framework is shown in Fig. \ref{Architecture}. Building upon classic semi-supervised learning (SSL) framework, we adopt SemiSAM, an additional supervision branch to utilize the segmentation output of SSL framework trained with domain knowledge to provide information for localization and generation of input prompts for SAM and the generated pseudo-labels of SAM as additional regularization to assist in the learning procedure of the semi-supervised framework.

\subsection{Semi-Supervised Framework}

Regarding the semi-supervised segmentation task, we denote the labeled set as $D_{L} = \{X_{i}, Y_{i}\}_{i=1}^{M}$ and the unlabeled set as $D_{U} = \{X_{j}\}_{j=1}^{N}$, where $X \in R^{H \times W \times D}$ and $Y \in \{0,1\}^{H \times W \times D}$ represent the input image and corresponding ground truth, and $M<<N$ in most scenarios. 
To utilize unlabeled data for semi-supervised learning, consistency regularization is a widely applied strategy by enforcing an invariance of predictions of input images under different perturbations and pushing the decision boundary to low-density regions, based on the assumptions that the perturbations should not change the output of the model.
The consistency regularization-based semi-supervised framework mainly consists of two components: \textbf{the main branch} with the segmentation network to input original images and generate main segmentation outputs, and \textbf{the consistency branch} to introduce perturbations to the input images or network conditions to generate additional segmentation outputs.
For labeled set $D_{L}$, we calculate supervised segmentation loss $\mathcal{L}_{seg}$ between the main segmentation outputs and ground truth. While for the unlabeled set $D_{U}$, we calculate the unsupervised consistency loss $\mathcal{L}_{con}$ between the main segmentation outputs and addition segmentation outputs, based on the assumption that the segmentation of the same image under different conditions should also be the same.

In the mean teacher (MT) framework \cite{tarvainen2017mean}, the main branch and the consistency branch are the student model and the teacher model, respectively. The teacher model is an average of student models over different training steps using exponential moving average (EMA) to ensemble weights of the student model as $ \theta^{'}_{t} = \alpha \theta^{'}_{t-1} + (1-\alpha) \theta_{t}$, where $\theta$ and $ \theta^{'}$ represent the weights of student model and teacher model, where $\alpha$ is a hyper-parameter named EMA decay.
Therefore, the mean-teacher framework can be formulated as training the network by minimizing the combination of supervised segmentation loss and unsupervised consistency loss as follows.

\begin{equation}
\min \limits_{\theta} \mathcal{L}_{sup}(f_{\theta}(X_{i}),Y_{i}) + \lambda_{c} \mathcal{L}_{con}(f_{\theta}(X_{j}),f_{\theta^{'}}(X_{j}))
\end{equation}

\subsection{SAM-Assisted Consistency Regularization}

Despite existing semi-supervised frameworks can utilize unlabeled data to improve the performance, when only one or a few labeled images are available, the model still cannot learn sufficient domain knowledge for accurate segmentation of challenging regions with complex structures, while the implementation of consistency learning may cause the model to generate "similar but incorrect" predictions of the majority of targets and ignore the inherent information.
Despite the limited performance, coarse segmentation can provide information for localization and generating prompt points for SAM.
Specifically, SAM is a large-scale pre-trained foundation model for promptable image segmentation tasks. Its architecture can be divided into three core components: an image encoder based on Vision Transformer (ViT)~\cite{ViT2020} to extract image embeddings, a prompt encoder to integrate user interactions via different prompt modes, and a lightweight mask decoder to predict segmentation masks by fusing image embeddings and prompt embeddings. 
To utilize SAM for assisting in semi-supervised medical image segmentation, we add an additional SAM-assisted supervision branch to generate relatively reliable pseudo-labels and guide the learning procedure.
To capture the spatial information of volumetric medical images, we utilize SAM-Med3D \cite{SAM-Med3D} in our framework by transforming the original 2D components into 3D counterparts.
Building upon existing semi-supervised frameworks, the outputs of the main branch are utilized as the proposals to generate input prompts for SAM as follows.

\begin{equation}
F_{\Theta}(X_{j}) = \mathbf{M}[ \mathbf{I}(X_{j}) + \mathbf{P}(f_{\theta}(X_{j})) ]
\end{equation}
where $\mathbf{I}$, $\mathbf{P}$, and $\mathbf{M}$ represent the image encoder, the prompt encoder, and the mask decoder, respectively.
To minimize the performance degradation caused by noisy prompts generated from coarse segmentation, we conduct an uncertainty-aware strategy to select out regions for the prompt generation to ensure more reliable guidance.
Since regions with high uncertainty mainly locate at challenging areas with ambiguous predictions, using prompt points generated from regions with low uncertainty will result in better performance compared with generating from the entire segmentation output. 

After that, we estimate the consistency loss $\mathcal{L}_{sam}$ between the main segmentation outputs and pseudo-labels generated from the SAM-assisted branch as additional supervision signals to assist in the semi-supervised learning procedure.
Therefore, the task can be formulated as training the network by minimizing the following functions.

\begin{equation}
\begin{split}
\min \limits_{\theta} \ & \mathcal{L}_{sup}(f_{\theta}(X_{i}),Y_{i}) + \lambda_{c} \mathcal{L}_{con}(f_{\theta}(X_{j}),f_{\theta^{'}}(X_{j})) + \\
 \ & \lambda_{s} \mathcal{L}_{con}(f_{\theta}(X_{j}),F_{\Theta}(X_{j}))
\end{split}
\end{equation}
where $\theta$, $\theta^{'}$, and $\Theta$ represent the weights of the student model, the teacher model and SAM.
To control the balance between supervised segmentation loss and consistency loss, following the design in related works \cite{yu2019uncertainty,zhang2021dual}, a ramp-up weighting coefficient $ \lambda_{c}=0.1*e^{-5(1-t/t_{max})}$ is used to mitigate the disturbance of consistency loss at early training stage, where $t$ represents the current number of iterations and $t_{max}$ represents the maximum number of iterations. 
For SAM consistency loss in our framework, we set a ramp-down weighting coefficient $ \lambda_{s}=0.1*e^{-5(t/t_{max})}$ to enhance the guiding of SAM at the early training stage and mitigate possible misleading in the late training stage.

\section{Experiments}

\subsection{Dataset and Implementation Details}

To evaluate the performance of the proposed method, we conduct experiments on the Left Atrium Segmentation Challenge Dataset \cite{LAdataset}.
The dataset contains 100 3D gadolinium-enhanced MR imaging scans (GE-MRIs) and corresponding LA segmentation masks for training and validation. These scans have an isotropic resolution of $0.625 \times 0.625 \times 0.625 mm^{3}$. 
All of our experiments are implemented in Python with PyTorch, using an NVIDIA A100 GPU. We use V-Net \cite{milletari2016v} as the semi-supervised backbone and SAM-Med-3D \cite{SAM-Med3D} as the SAM backbone for segmentation of 3D volumetric medical images. 
Following the same task setting in \cite{yu2019uncertainty}, we divide the 100 scans into the same 80 scans for training and 20 scans for testing, and apply the same implementation details like optimizer, iterations, and learning rate decay, excluding the patch size changed from $112 \times 112 \times 80$ to $128 \times 128 \times 128$ sub-volumes to fit in the input of SAM-Med3D \cite{SAM-Med3D}. We use the SGD optimizer to update the network parameters with an initial learning rate of 0.01 and divided by 10 every 2500 iterations, with the maximum number of iterations as 6000. We use four commonly used metrics in segmentation tasks for evaluation including the Dice similarity coefficient (Dice), Jaccard Index (Jaccard), 95\% Hausdorff Distance (95HD), and Average Surface Distance (ASD). Lower 95HD and ASD indicate better segmentation performance, while larger Dice and Jaccard indicate better segmentation results. 
Our code will be available at https://github.com/YichiZhang98/SemiSAM.

\begin{table}[t]
	\caption{Comparison of segmentation performance with training-based automatic segmentation methods and SAM-based interactive segmentation methods on Left Atrium segmentation dataset. To adapt 2D SAM to 3D medical images, N denotes the count of slices containing the target object ($N \approx 60$ for LA dataset).} \label{Table1}
	\centering
    \scriptsize
    \setlength\tabcolsep{3pt}
	\renewcommand\arraystretch{1.6}
	\begin{tabular}{c|c|c|c|c}
		\hline 	\hline
		\bf{Method}  & \textbf{Manual Annotation}  & \textbf{Manual Prompt} &  \textbf{Dice$\uparrow$[\%]} & \textbf{Jaccard$\uparrow$[\%]} \\ \hline
            \multirow{4}{*}{FS Baseline}   & 1 labeled case    & \multirow{4}{*}{-}             & 17.06     & 12.25                    \\
               & 2 labeled cases            &                     & 40.48     & 26.43   \\
               & 4 labeled cases            &                     & 54.43     & 39.60   \\
               & 8 labeled cases            &                     & 76.09     & 64.12                    \\ \hline
            \multirow{4}{*}{SAM  \cite{SAM-Meta}}         & \multirow{4}{*}{-}                         & N points             & 41.98    & 27.00      \\
                     &                          & 2N points           & 46.11    & 30.41       \\
                     &                          & 5N points           & 58.90    & 42.32       \\
                     &                          & 10N points           & 71.38    & 56.03       \\ \hline
		    \multirow{4}{*}{SAM-Med3D \cite{SAM-Med3D}}   & \multirow{4}{*}{-}              & 1 points             & 56.51    & 40.74          \\
               &                 & 2 points             & 63.96    & 48.47          \\
               &                 & 5 points             & 70.28    & 55.13          \\
               &                 & 10 points            & 73.78    & 59.27          \\  \hline
            \multirow{3}{*}{MT \cite{tarvainen2017mean}}     & 1 labeled case     & \multirow{3}{*}{-}          & 30.60     & 19.82            \\
                 & 2 labeled cases            &                 & 58.17     & 44.60            \\ 
                 & 4 labeled cases            &                 & 72.40     & 59.19            \\ \hline
            \multirow{3}{*}{\textbf{SemiSAM-MT}}                    & 1 labeled case       & \multirow{3}{*}{-}                     & 41.38 & 27.95  \\
                        & 2 labeled cases            &                   & 69.46 & 55.26 \\  
                        & 4 labeled cases            &                   & 80.42 & 68.05 \\ \hline
                        
            FS Upperbound   & 80 labeled cases          & -                    & 91.14     & 83.82                    \\  \hline \hline
	\end{tabular}
\end{table}

\subsection{Comparison with Other Segmentation Methods}

In this section, we make a comparative analysis between training-based automatic segmentation methods and SAM-based interactive segmentation methods on the Left Atrium segmentation dataset. Specifically, since the original SAM \cite{SAM-Meta} is based on 2D images for segmentation, the prompt should be given on each slice containing the target object to obtain the segmentation of the whole 3D image.
Table. \ref{Table1} presents the segmentation performance of different methods on the test set.
For SAM-based interactive segmentation methods, it can be observed that an increase in the number of prompt points leads to improved segmentation performance. Additionally, by utilizing volumetric information, SAM-Med3D demonstrates superior performance over classic SAM with significantly fewer prompt points. 
Although these methods can achieve zero-shot segmentation without the need for pixel/voxel-level annotated data for training, high manual efforts are still needed for labeling sufficient prompt points of each testing image to obtain acceptable performance. 
For training-based automatic segmentation methods, the segmentation performance increases as the number of labeled training data increases. However, when only one or a few labeled images are available, the performance is far behind fully supervised performances. 
Our method can make use of SAM as an additional supervision signal without the need for manual prompting to assist in the learning procedure and further improve the performance of existing semi-supervised frameworks with 10.78 \%, 11.29 \%, and  8.02 \% in dice similarity coefficient using only 1, 2, and 4 labeled cases.

Table. \ref{Table2} shows the performance of our method with comparison to other semi-supervised frameworks using different numbers of labeled images.
Building upon the vanilla mean teacher framework, utilizing SemiSAM significantly improves the performance and outperforms other cutting-edge semi-supervised medical image segmentation methods in scenarios with extremely limited annotations.
However, we observe that when a relatively large amount (e.g. 10\%) of labeled data is available, simply enforcing the consistency of outputs between SAM and semi-supervised segmentation model may not significantly improve the segmentation performance, since the performance of training-based segmentation methods exceeds SAM-based zero-shot segmentation methods in such scenarios.

\begin{table}[]
	\caption{Quantitative comparison of adapting SemiSAM upon mean teacher (MT) with fully-supervised other semi-supervised segmentation methods on LA dataset with different number of annotation cases. The arrows of evaluation metrics indicate which direction is better.} \label{Table2}
	\centering
    \setlength\tabcolsep{3pt}
    \scriptsize
	\renewcommand\arraystretch{1.6}
	\begin{tabular}{c|c|c|c|c|c}
		\hline  \hline
		\textbf{Method}	& \textbf{Label/Unlabel}	& \textbf{Dice$\uparrow$[\%]} & \textbf{Jaccard$\uparrow$[\%]} & \textbf{ASD$\downarrow$[vol]} & \textbf{95HD$\downarrow$[vol]}  \\ \hline
	Supervised                   &  1/0       & 17.06 & 12.25 & 23.61 & 65.17       \\ \hline
          	UA-MT  \cite{yu2019uncertainty}   &  1/79    & 31.66 & 20.03 & 24.66 & 50.43 \\   
           Ent-Mini  \cite{vu2019advent}      & 1/79    & 31.59 & 19.43 & 26.24 & 51.04 \\
    DTC  \cite{luo2021semi}      &  1/79     &  36.94 & 23.29 & 23.88 & 52.53     \\ 
    DTML \cite{zhang2021dual}  &  1/79  & 37.31 & 22.93 & 24.22 & 47.70   \\ \hline
         MT  \cite{tarvainen2017mean}      &  1/79     &  30.60 & 19.82 & 25.98 & 51.89  \\ 
         \textbf{+SemiSAM} &  1/79    & 41.38 & 27.95 & 20.60 & 45.79  \\  \hline \hline
    
	Supervised                   &  2/0        &  40.48 & 26.43 & 17.16 & 42.09      \\ \hline
      	UA-MT  \cite{yu2019uncertainty}   &  2/78    & 59.76 & 44.77 & 15.28 & 40.58  \\
       Ent-Mini  \cite{vu2019advent}      & 2/78    & 66.02 & 50.36 & 13.46 & 36.44 \\
    DTC  \cite{luo2021semi}      &  2/78     & 65.37 & 48.56 & 9.84 & 31.32 \\
    DTML \cite{zhang2021dual}  &  2/78  & 67.80 & 53.52 & 6.44 & 22.81 \\  \hline
          MT  \cite{tarvainen2017mean}      &  2/78     & 58.17 & 44.60 & 17.33 & 44.93   \\ 
     \textbf{+SemiSAM} &  2/78    & 69.46 & 55.26 & 10.65 & 31.67  \\  \hline  \hline
    
	Supervised                  &  4/0        &  54.43 & 39.60 & 9.87 & 47.05     \\ \hline
      	UA-MT  \cite{yu2019uncertainty}   &  4/76    &  73.46 & 60.12 & 4.71 & 17.41  \\
       Ent-Mini  \cite{vu2019advent}      &  4/76     & 77.61 & 64.63 & 7.44 & 23.51         \\ 
    DTC  \cite{luo2021semi}      &  4/76     & 78.04 & 65.18 & 7.65 & 25.33         \\ 
    DTML \cite{zhang2021dual}  &  4/76  &  79.46 & 67.25 & 7.04 & 24.81 \\ \hline
          MT  \cite{tarvainen2017mean}      &  4/76   &  72.40 & 59.19 & 6.71 & 21.55  \\ 
        \textbf{+SemiSAM} &  4/76     &  80.42 & 68.05 & 5.16 & 18.23 \\  \hline  \hline
    
	Supervised &  8/0       & 76.09 & 64.12 & 6.60 & 24.04    \\  \hline
         	UA-MT  \cite{yu2019uncertainty}   & 8/72     & 84.25 & 73.48  & 3.36 & 13.84   \\ 
          Ent-Mini  \cite{vu2019advent}      & 8/72     & 86.43 & 76.30  & 2.94 & 12.86\\
    DTC \cite{luo2021semi}   & 8/72   & 86.57    &76.55  &3.74   &14.47            \\ 
    DTML \cite{zhang2021dual}  &  8/72  & 87.13 & 77.49 & 3.25 & 13.81  \\ \hline 
          MT  \cite{tarvainen2017mean}      &  8/72  &  83.09 & 71.88 & 5.14 & 17.63  \\ 
          \textbf{+SemiSAM} &  8/72    & 84.45 & 73.75 & 3.31 & 14.56 \\  \hline\hline
 Supervised                   &  80/0             &  91.14  & 83.82 & 1.52 & 5.75   \\ \hline  \hline
	\end{tabular}
\end{table}

\begin{figure}[t]
	\includegraphics[width=9cm]{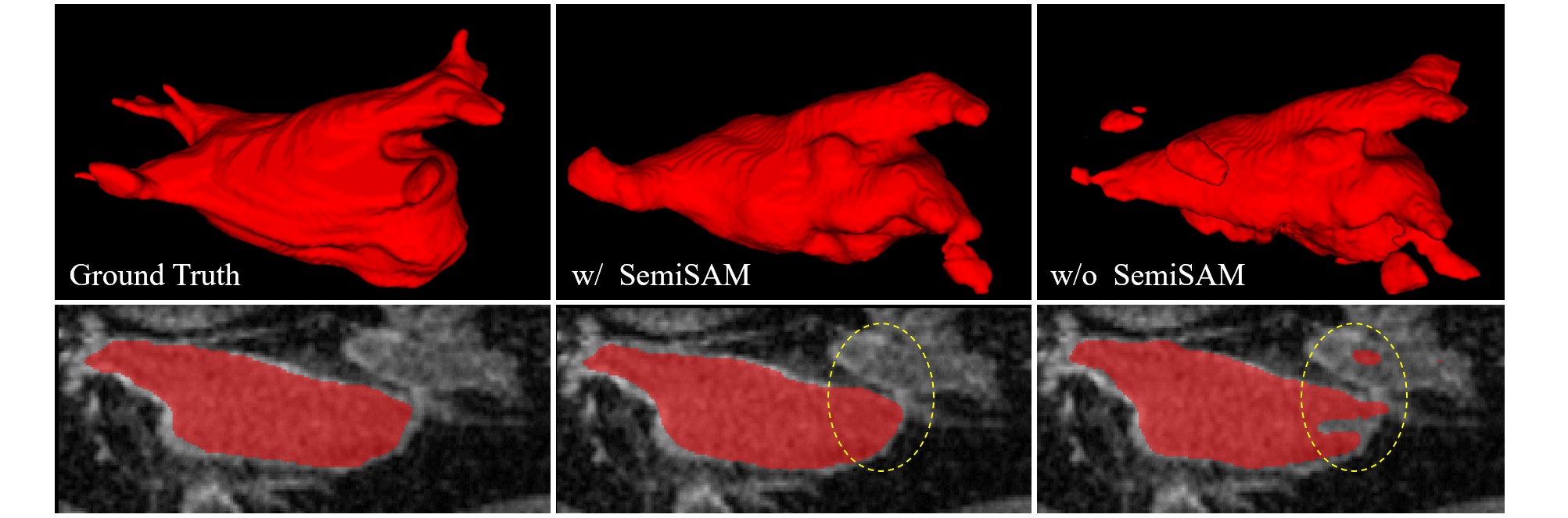}
	\caption{Visual comparison of 3D reconstruction results and 2D segmentation results of semi-supervised mean teacher frameworks with (w/) and without (w/o) SemiSAM on LA dataset.}
\end{figure}

\section{Conclusion and Discussion}

In this paper, we propose a simple yet efficient strategy to explore the usage of SAM as an additional supervision branch for enhancing consistency regularization-based semi-supervised medical image segmentation framework in scenarios with extremely limited annotations.
Instead of classic supervised segmentation loss based on labeled cases and unsupervised consistency loss based on unlabeled cases, we leverage additional regularization of predictions between SAM and the semi-supervised model as an additional supervision signal to assist in the learning procedure.
Experimental results demonstrate that SemiSAM further improves the segmentation performance of semi-supervised frameworks especially under extreme few-annotation settings where only one or a few labeled data are available.
Our work provides new insights for medical image segmentation where acquiring labeled data is difficult and expensive by leveraging the knowledge base of foundation models. 

While our method has demonstrated significant improvements, there are aspects where further enhancements could be made. 
Firstly, we only conduct relatively simple experiments on one representative SSL benchmark to present the motivation and effectiveness of proposed method.
As a general plug-and-play framework, SemiSAM could be easily adapted to other 2D/3D SAM variants \cite{SAM2-Meta,SAM2-MIS} and applied to more up-to-date SSL frameworks for other segmentation tasks without additional modifications.
Besides, the improvement of simply enforcing the consistency is limited when a relatively large amount of labeled data is available.
In the future, we intend to explore more robust and reliable utilization of SAM to ensure consistent improvement and further extend the flexibility of SemiSAM as a plug-and-play strategy to fit in more semi-supervised frameworks with more comprehensive evaluations.

\section*{Acknowledgement}
This paper was supported by the National Natural Science Foundation of China under Grant 82394432 and 92249302, and Shanghai Municipal Science and Technology Major Project under Grant 2023SHZDZX02. The computations in this research were performed using the CFFF platform of Fudan University.

\bibliographystyle{IEEEtran}
\bibliography{ref}

\end{document}